\documentclass[a4paper,twoside]{article}

\usepackage{epsfig}
\usepackage{subfigure}
\usepackage{calc}
\usepackage{amssymb}
\usepackage{amstext}
\usepackage{amsmath}
\usepackage{amsthm}
\usepackage{multicol}
\usepackage{pslatex}
\usepackage{apalike}
\usepackage{graphicx}
\usepackage{color}
\usepackage{balance}
\usepackage{SCITEPRESS}     

\begin{document}

\title{Phase Distribution in Probabilistic Movement Primitives, Representing Time Variability for the Recognition and Reproduction of Human Movements}

\author{\authorname{Vittorio Lippi, Raphael Deimel}
\affiliation{Technische Universit{\"a}t Berlin, Fachgebiet Regelungssysteme, Sekretariat EN11, Einsteinufer 17, -10587 Berlin, Germany}
\email{\{vittorio.lippi@tu-berlin.de, raphael.deimel@tu-berlin.de}
}


\abstract{Probabilistic Movement Primitives (ProMPs) are a widely used representation of movements for human-robot interaction. They also facilitate the factorization of temporal and spatial structure of movements. In this work we investigate a method to temporally align observations so that when learning ProMPs, information in the spatial structure of the observed motion is maximized while maintaining a smooth phase velocity. We apply the method on recordings of hand trajectories in a two-dimensional reaching task. A system for simultaneous recognition of movement and phase is proposed and performance of movement recognition and movement reproduction is discussed.}

\onecolumn \maketitle \normalsize \vfill

\section{\uppercase{Introduction}}
\paragraph{Overview.} Probabilistic movement primitives (ProMP) \cite{paraschos2013probabilistic} are a representation of movements used in robot control and human-robot interaction (HRI) applications~\cite{maeda2017probabilistic} and provide several desirable properties to model tasks for robot control and HRI\cite{paraschos2018using}. One of these properties is temporal scaling: the movement trajectory is not a direct function of time but function of a \textit{phase} variable, $\phi(t)$, i.e. the temporal evolution of $\phi$ determines the the velocity of the movement independently of the spatial structure. 

\begin{figure}[htbp]
	\centering
		\includegraphics[width=1.00\columnwidth]{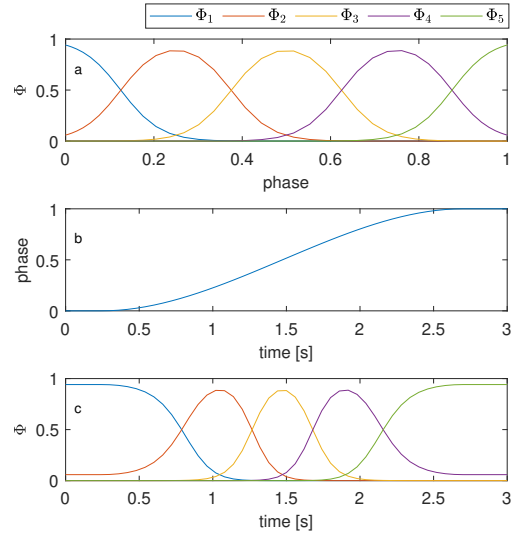}
	\caption{Features and Phase. In (a) the features $\Phi_i(\phi)$ are represented as a function of the phase. In (b) a phase profile, based on the \textsl{beta function} as a function of time. In (c) the features are represented as a function of the time.}
	\label{fig:Fig1BasisNphase}
\end{figure}

A ProMP represents a sample movement $y(t)$ as
\begin{equation}
	y(t)=\left[ 
	\begin{array}[pos]{c}
		q \\
		\dot{q}
	\end{array}\right]
	=\Phi(\phi(t))w+\epsilon
	\label{eqdef}
\end{equation}
where $q$ is the vector of variables describing the movement (usually joint angles or hand effector position and orientation) and $\Phi$ is a $Nx2$ vector computed with Gaussian functions and their derivatives:
\begin{equation}
	\Phi_i = \frac{e^{(\phi(t)-c_i)^2/h_i}}{\sum^{N}_{k=1} e^{(\phi(t)-c_k)^2/h_k}}
	\label{feateq}
\end{equation}
where $c_i$ represents the \textit{center} and $h_i$ expresses the spread of the bell-shaped feature. The vector $w$ is drawn from a multinomial distribution which is defined by its parameters $\theta$: 
\begin{equation}
p(w|\theta)=\mathcal{N}(\mu_w,\Sigma_w)
	\label{wdist}
\end{equation}
In a general formulation, see for example \cite{colome2014dimensionality} the $\epsilon$ in Eq. \ref{eqdef}  is characterized by a distribution such as $p(\epsilon)=\mathcal{N}(0,\Sigma_y)$. In this paper we assume that all the variability observed on $y$ should be accounted by the distribution of $w$ and that $\epsilon \approx 0$. In this way each observed trajectory $y_i$ is represented by a single weight vector $w_i$. Vice versa, $w$ can be interpreted as a compressed representation of the movement trajectory, obtained from projecting trajectory values onto a lower dimensional subspace using the Moore-Penrose pseudo-inverse of $\Phi$:
\begin{equation}
	w=\Phi^+ y
	\label{eqw}
\end{equation}
Which weight vector is most likely depends on the optimization criterium, e.g. jerk minimization as regularization principle \cite{paraschos2018using}. Alternative features for movement representation can be designed to guarantee other properties such as minimal error at the final position and speed \cite{lippi2012method}. Due to the probabilistic nature of the ProMP framework constrains on generated trajectories can be imposed by conditioning the probability distribution\cite{paraschos2018using}. 
Figure \ref{fig:Fig1BasisNphase} illustrates the features as either a function of time or phase. The time modulation performed by the phase function also modulates the resolution of the low dimensional representation during movement execution. In Figure \ref{fig:Fig1BasisNphase}~(c) features in the middle of the movement are more dense than at the beginning or at the end, due to the choice of a sigmoidal phase profile. We assume that part of the variability observed among the repetitions of a movement is caused by shifted and stretched phase signals. Time scaling can be performed by means of a linear constant, i.e. $\phi= \alpha t$ \cite{maeda2017probabilistic,ewerton2015modeling}. In this work we estimate phase profile parameters for each observation that minimize the variances over all observed $w$ (or, equivalently of the response in phase domain), under the constraint of an assumed structure for the phase profile. 
In particular we propose to restrict phase signals to the beta function:
\begin{equation}
\phi_{\Delta_1,\Delta_2}(t)=\int_0^t \beta_{2,2}\left(\frac{\tau-\Delta_1}{\Delta_1-\Delta_2}\right) d\tau
\label{betaeq}
\end{equation}
where $\beta_{2,2}$ is the incomplete beta function with parameters $a=b=2$. This sigmoidal function, shown in \ref{fig:Fig1BasisNphase} (b), is differentiable, monotonic and saturates at $0$ and $1$. The parameters $\Delta_1$, $\Delta_2$ associated with each trajectory depend on the distribution of the whole sample set. Notice that weights $w$ are in general specific w.r.t.\ to the chosen phase profile (see Eq. \ref{eqw}). The average of the sample set distribution is defined as:
\begin{equation}
	\overline{y}= \underset{\widetilde{y}}{argmin} \sum_{i=1}^{N}\int_{0}^{1}\left(y_i(\phi)-\widetilde{y}\left(\phi\right)\right)^2d\phi
	\label{mina}
\end{equation}
For the $i^{th}$ sample the phase is defined as: 
\begin{equation}
	\phi_i = \underset{\widetilde{\phi}}{argmin} \int_{0}^{1}\left(y_i\left(\widetilde{\phi}\left(\tau(\phi)\right)\right)-\overline{y}(\phi)\right)^2d\phi
  \label{minb}
\end{equation}
The term $\tau(\phi)$ represents alignment between the average $\overline{y}$ and the $y_i$ sample, it express the fact that as $\widetilde{\phi}$ and $\phi$ are both monotonically dependent on $t$, the former can be expressed as a function of the latter, in particular:   
\begin{equation}
  \tau(\phi)=arg_t \left(\phi(t)=\widetilde{\phi}(t)\right)
\end{equation}

\begin{figure*}[htb]
	\centering
	 \begin{tabular}{cc}

		\includegraphics[width=0.90\columnwidth]{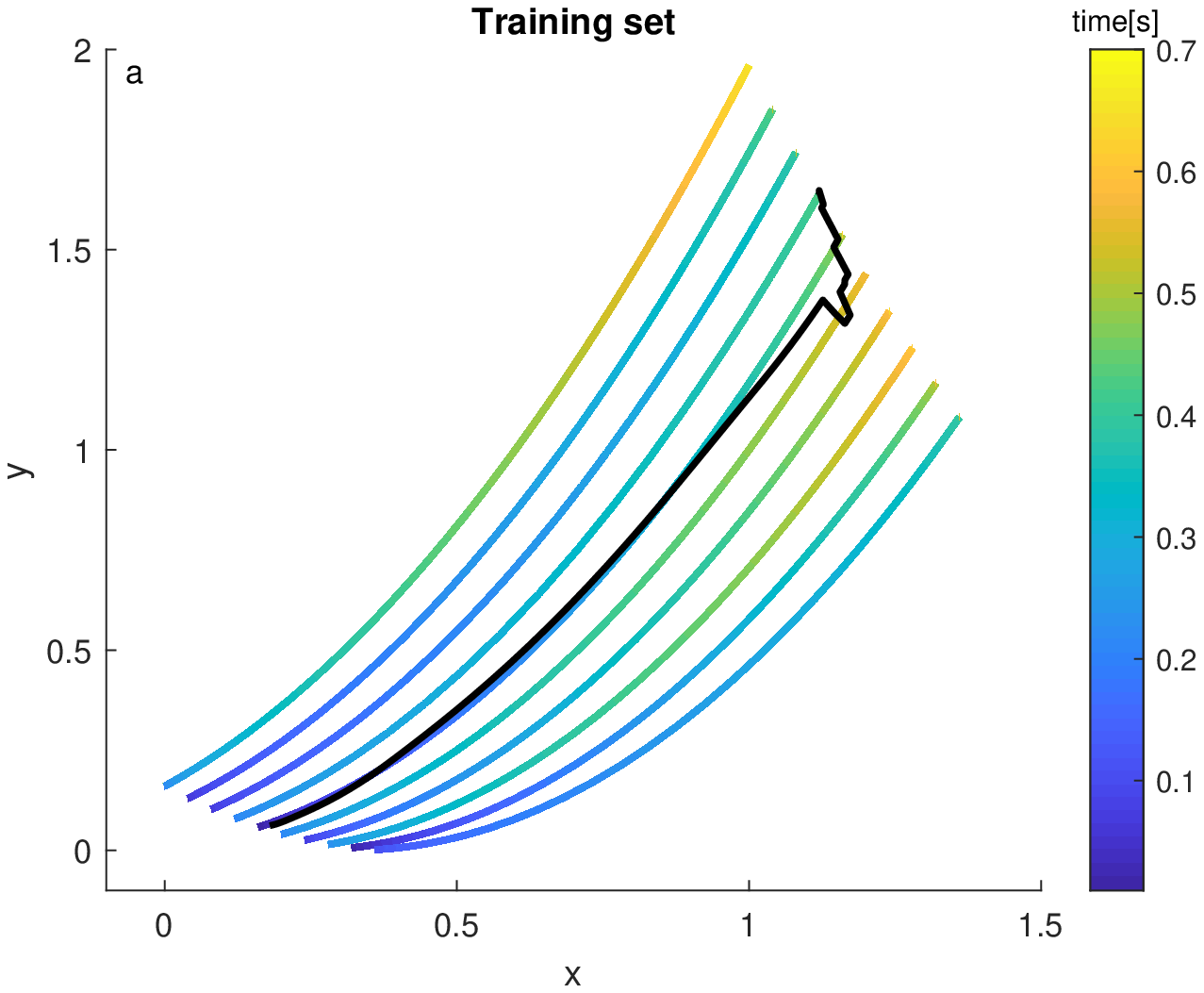} &
		\includegraphics[width=0.90\columnwidth]{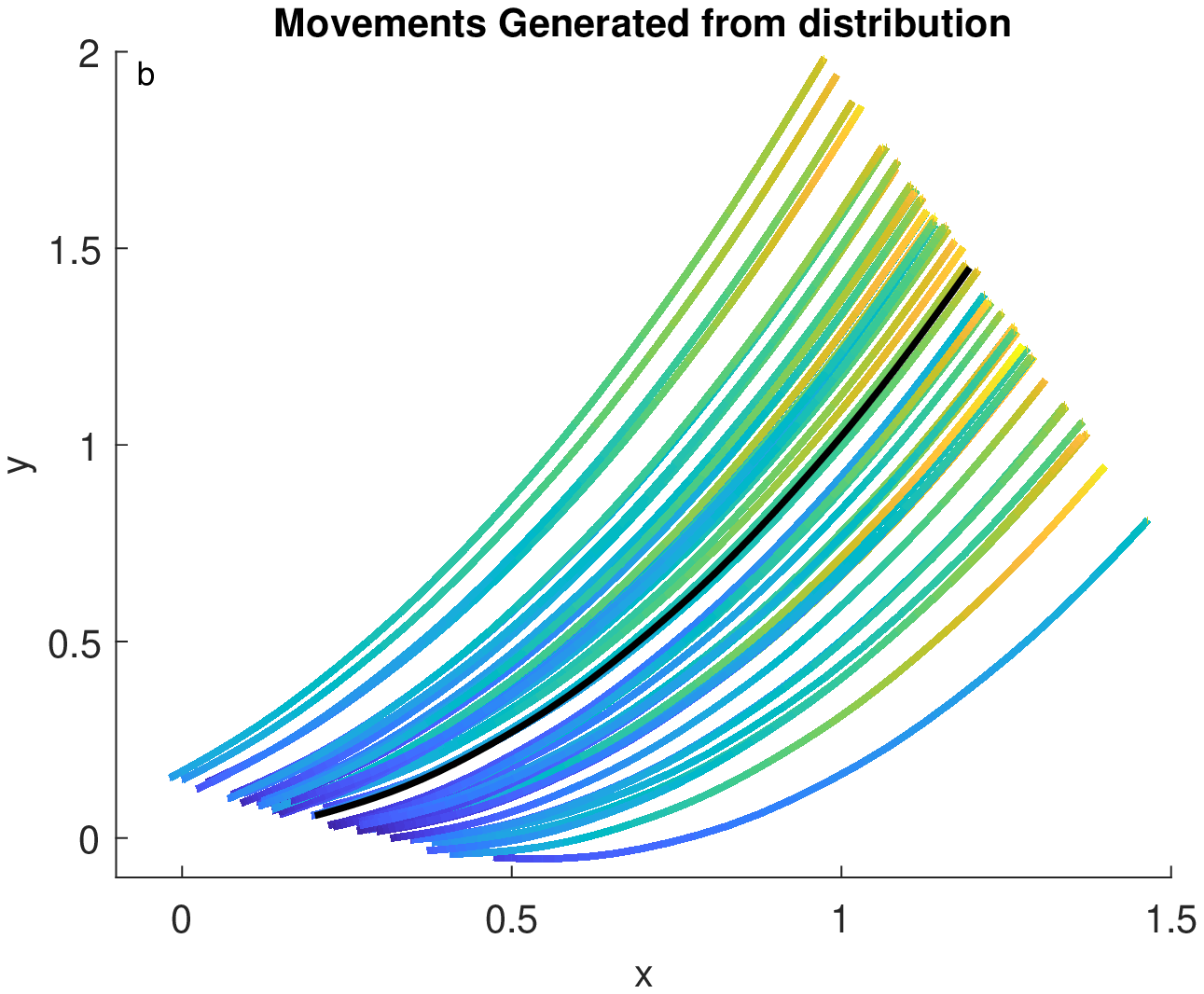} \\
		\includegraphics[width=0.90\columnwidth]{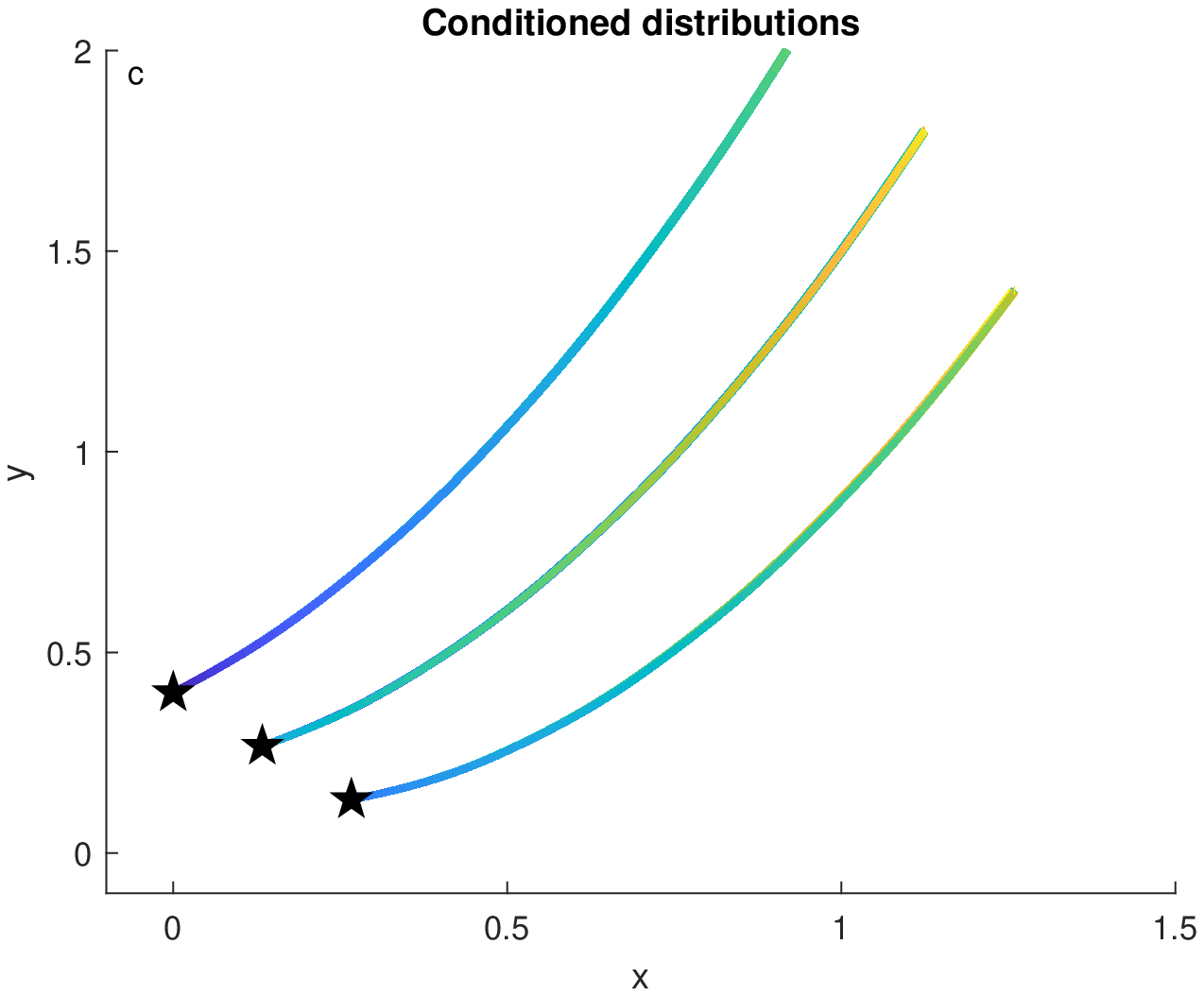} &
		\includegraphics[width=0.90\columnwidth]{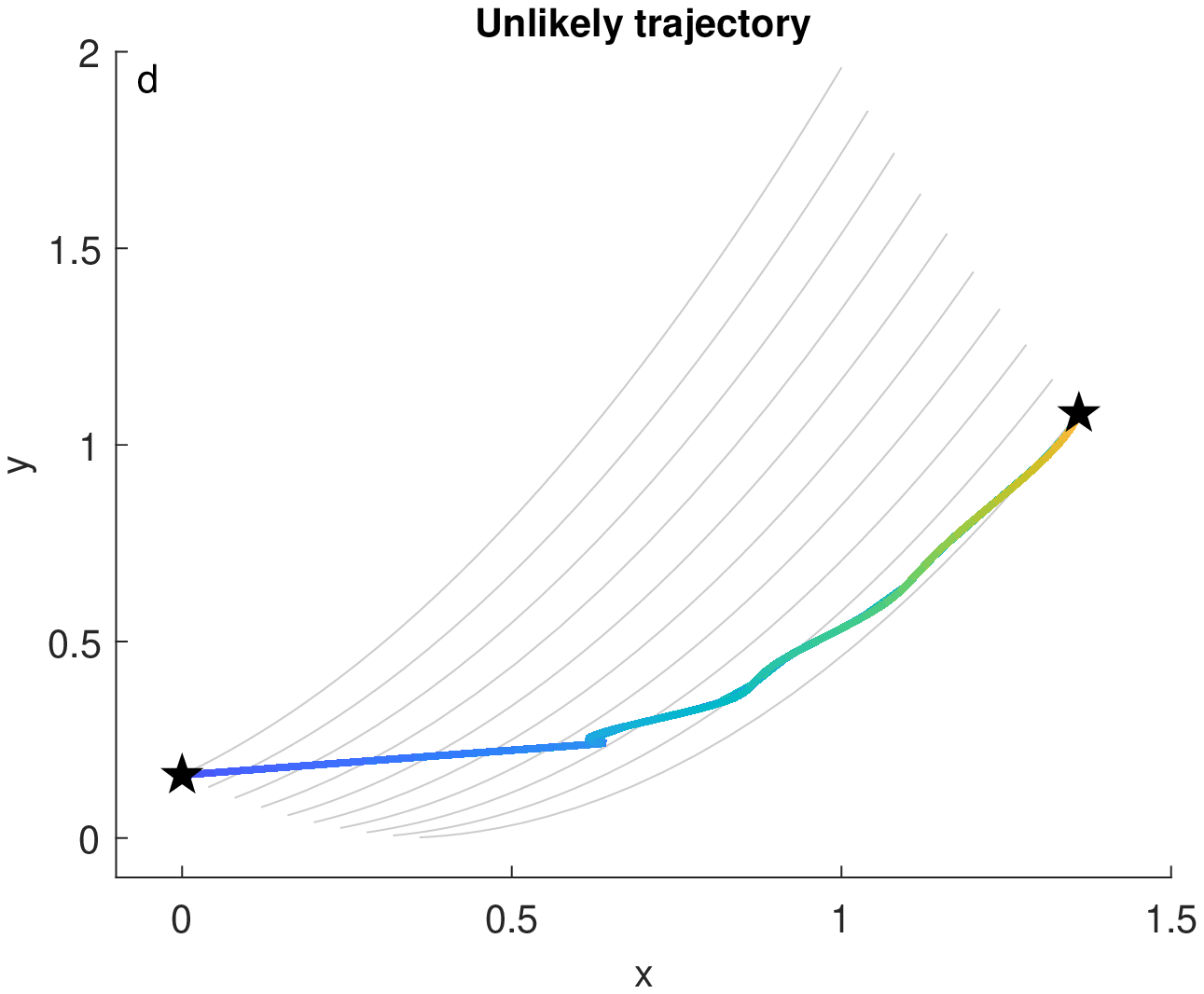} \\
		
		\end{tabular}
	\caption{Parabolic lines synthetic data set. In (a) an artificial data-set of trajectories with a parabolic shape and random duration is shown, with only 10 of the 100 produced samples shown but covering more intermediate positions. The color gradient indicates the temporal evolution of the trajectory, the black line the average trajectory in time. The resulting shape does not match the presented examples well, especially in the end when sample count drops. The data-set has been used as a training set for a proMP model. In (b) trajectories generated with the proMP distribution are shown, the black line represent the average (computed respect to phase). Notice that while the model captured very accurately the shape of the movements, the gaussian distribution of parameters results in a distribution of trajectory that is different from the one presented in the training set. In (c) conditioned trajectories with imposed via-points a time $t=0$ are shown, indicated by black stars. The conditioned $\Sigma_w$ is so small that the trajectories generated for each via-point are indistinguishable. In (d) the trajectory is conditioned to pass through two via-points unlikely to appear in the same movement. The resulting shape is dissimilar to those presented in the training set. For reference, trajectories from the training set are shown in gray.}
	\label{lineexample}
\end{figure*}

As $\overline{y}$ depends on all phase profiles, Eq. \ref{mina} and Eq.\ref{minb} need to be optimized together to find the optimal phase profiles. The problem is solved by iteratively optimizing $\overline{\phi}(t)$ and $\phi_i$, which yields phase profile parameters for each sample movement. The phase profile parameters of the average trajectory is fixed a priori to $\Delta_1 =\Delta_2 = 0$, as the time scaling when comparing two or more movements (e.g. when computing the average) depends on the relationship between the respective phase profiles and hence there is a degree of redundancy. 
The described time scaling can also be performed on time domain values. In order to take into account the movement primitive representation the $y$ in Eq. \ref{mina} and Eq. \ref{minb} can be projected on the MP representation $y_{mp}=\Phi w$.
Once optimized time scaling parameters of the sample set are obtained, we can obtain a probability distribution describing the movements using empirical estimators from literature~\cite{maeda2017probabilistic,paraschos2018using}. In the following paragraphs we demonstrate (a) how to identify a model for movements, (b) how to recognize an observed movement given a set of movement primitives, (c) how to estimate the current phase, (d) how to integrate perception in the phase recognition process and (e) how to generate a movement.
\paragraph{Model identification.} The estimation of ProMP parameters from of a sample set can be based on different principles, e.g. linear regression of each observation individually~\cite{paraschos2013probabilistic} or maximizing the likelihood of the complete observed data\cite{paraschos2018using}. As introduced in the overview in this work we assume $\epsilon \approx 0$ in Eq. \ref{eqw} and that the phase profile can be chosen to minimize the variability on $w$. This leads to a two step procedure: First, each observed trajectory is assigned a set of parameters $(w_i, \Delta_{1i}, \Delta_{2i})$.  
This is done on the basis of Eq. \ref{mina} and Eq. \ref{minb} to obtain the phase profile, and Eq. \ref{eqw} to obtain $w$, then the distributions for $w$ (Eq. \ref{feateq}) and $\phi(t)$ is obtained with an empirical estimator. 
An example of model identification is shown in Fig.\ref{lineexample}. The training set is generated artificially and consists of planar motion with random time profile (with the form specified in Eq.\ref{betaeq}) and non-intersecting parabolic trajectories. In Fig.\ref{lineexample} (b) movements generated with the identified distribution are shown. Notice that assuming a normal distribution of $w$ produces a distribution of trajectories that is different from the more ``uniform'' training set.

\paragraph{Movement recognition} An observed movement $y$ can be classified by comparing likelihoods between ProMPs. The likelihood
\begin{equation}
\begin{array}[h]{l}
	p\left(y(t_i) | \mu_k, \Sigma_k \right) \\
	=\int_0^1 p\left(y(t_i)\,|\,\Phi(\phi) \mu_k,\Phi(\phi) \Sigma_k \Phi(\phi)^T\right) p(\phi|t_i) d \phi
	\label{movid}
	\end{array}
\end{equation}
of observing the movement given model $k$ is computed and the most likely model is selected, i.e. $L = \underset{k}{argmax} \, p(y|\theta_k)$. 
Notice that the probability of $y(t_i)$ is considered independent of the previous observations, i.e. $p(y(t_i)|\theta_k,y(t_{i-1})) = p(y(t_i)|\theta_k)$. The probability of an observed sequence can be obtained by multiplying the probabilities of the observed samples, $p(y|\theta_k) = \prod_{0}^{t_f} p(y_t|\theta_k)$.
In the described examples we will include all the past observations of $y_t$, in the general case not all points of the trajectory may be available, due to occlusion for example. Applying the ProMP framework to such cases is demonstrated in \cite{maeda2017probabilistic}.

\paragraph{Phase recognition} Using the Bayes rule we can estimate the most likely phase of an observed sample: 
\begin{equation}
p(\phi|y(t_i))=\frac{p(y(t_i)|\Phi(\phi)\mu_k,\Phi(\phi) \Sigma_k \Phi(\phi)^T) p(\phi|t_i)}{\sum_k^M p(\theta_k) p(\phi|y(t),\theta_k)}
\label{phaseid}
\end{equation}
Notice that $p(\phi|t_i)$ is assumed to be normally distributed and hence parameterized by mean $\mu_{\phi}(t)$ and standard deviation $\sigma_{\phi}(t)$ as functions over time. $\Delta_1$ and $\Delta_2$ are not normally distributed due to the nonlinear relationship in Eq. \ref{betaeq}. In most prior work on the topic, $\phi$ is assumed to be a linear function over time and that therefore $\phi(t)$ can be described by a normal distribution, e.g. in \cite{ewerton2015modeling}. Dynamic time warping (DTW) is another, nonlinear, approach proposed recently \cite{ewerton2018assisting}.

\paragraph{Perception.}
Eq.~\ref{phaseid} provides an estimate given a time-dependent but otherwise fixed $p(\phi|t_i)$ derived from the training set. Alternatively, we can use an explicit phase estimate provided by some perceptual system. 
Examples of fusing ProMP and perception have been proposed, e.g.\ in \cite{dermy2019multi} ProMPs are used in the context of predicting human intentions. In general the sensor input will have many more dimensions than $y$. In this work we will estimate the phase from recent observations of the motion. We will consider a phase estimator based on previous values of $y$ over a sliding time window to yield an estimate $\phi^*(t_i)$ of the current phase. Such estimator is used to compute the probability of the observed $y(t_i)$. Instead of integrating over the distribution of probability of $\phi$ as shown in Eq. \ref{movid}, only the estimated value is used to compute $y(t_i)$:
\begin{equation}
\begin{array}[h]{l}
	p(y(t_i) | \mu_k, \Sigma_k) \\
	= p \left( y(t_i)\,|\,\Phi(\phi^*(t_i)) \mu_k, \Phi(\phi^*(t_i)) \Sigma_k \Phi(\phi^*(t_i))^T \right)
	\end{array}
	\label{percphase}
\end{equation} 
\paragraph{Movement Generation.} The model can be used to generate movements. Depending on the task the movements can be generated deterministically using Eq. \ref{eqdef} or stochastically by sampling $w$ from the distribution. In certain cases the term $\epsilon$ in Eq. \ref{eqdef} may be nonzero due to actuation and external noise, but is not the case in the presented example. Generating movements from ProMP can be used both for robot control and to predict human movements in HRI, see, for example, \cite{oguz2017progressive}. Samples can obey constraints such as a specific target position or velocity, by conditioning the distribution to the constraint prior to sampling~\cite{paraschos2013probabilistic}, which can be done parameterically (and therefore efficiently) with normal distributions. A constraint can be expressed as a target point $y^*(\phi_i)$ to be reached at a given phase. Exploiting the linear dependency between $w$ and $y$ the updated parameters for $p(w|y_{\phi_i}=y^*(\phi_i))$ are
\begin{equation}
	\mu^*_{w}=\mu_w +\Sigma_w \Phi^T\!(\phi) \left( \Phi(\phi)\Sigma_w \Phi^T\!(\phi)\right)^{-1} \left(y^*-\Phi(\phi)\mu_w\right)
	 \label{munew}
\end{equation} 

 \begin{equation}
	\Sigma^*_{w}=\Sigma_w +\Sigma_w \Phi^T\!(\phi) \left( \Phi(\phi)\Sigma_w \Phi^T\!(\phi)\right)^{-1} \Phi(\phi)\Sigma_w
	\label{sigmanew}
\end{equation} 

An example of generated movements is shown in Fig.\ref{lineexample} (b,c,d). The conditioned distributions in \ref{lineexample}~(c) exhibit very small $\Sigma_w$ (i.e. all the trajectories starting from a given point are indistinguishable in shape), capturing the regularity of the training set. In Fig.\ref{lineexample} (d) the distribution is forced to pass through two via-points that are unlikely to be observed in the same movement, i.e. it is conditioned on an outlier. As a consequence, the resulting sampled trajectories do not resemble the observations presented in the training set.  

\section{Dataset and Benchmark Problem}

\begin{figure}
	\centering
		\includegraphics[width=1.00\columnwidth]{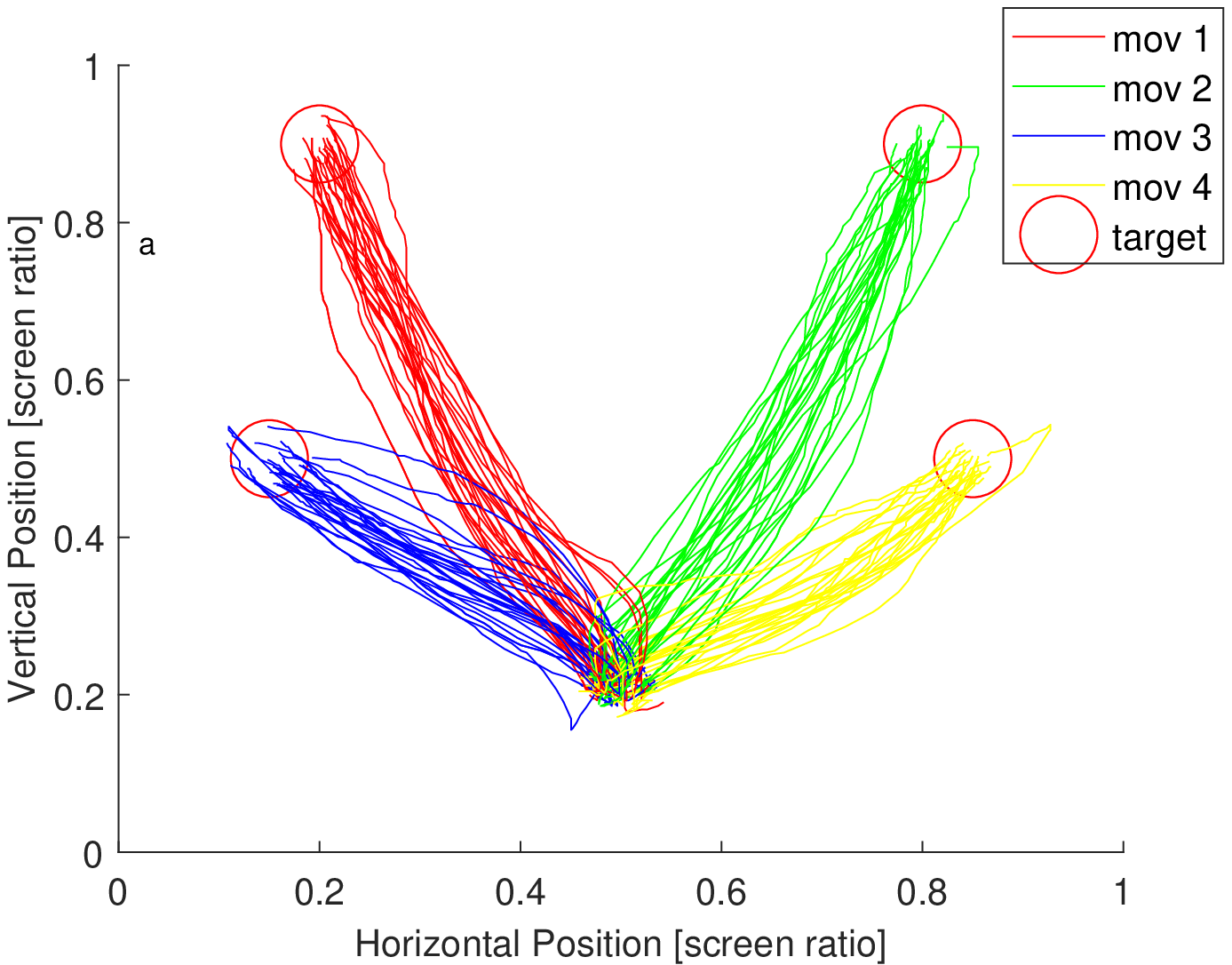}
	\caption{Reaching data set. A two-dimensional reaching task is performed by a test participant on a computer screen using a mouse. The participant was asked to perform the movement as fast as possible to reach circular targets that were appearing in a position randomly decided among $4$. The resulting set is composed by $25$ repetitions of $4$ distinct targets.}
	\label{dataset}
\end{figure}

The Reaching data set consists in a reaching task: Participants have been asked to move the mouse cursor from a starting point to targets appearing in 4 different positions. Each target identifies a different movement, associated to a different ProMP model. The samples are shown in Fig. \ref{dataset}. Given the nature of the task, consisting in reaching movements, it comes natural to perform the segmentation on the basis of events, in particular: (a) a movement starts when the mouse cursor is in the starting position and the target is shown on the screen and (b) it stops when a target is reached. In general different heuristics are possible for segmentation, notably the ProMP framework provides the possibility to segment the signal exploiting an \textit{expectation-maximization} algorithm, treating segmentation as a latent variable to be optimized together with the movement models parameters\cite{lioutikov2017learning}. The dataset includes 100 movement samples used as training-set for the \textit{model identification} and $100$ samples used as test set for the \textit{movement recognition} and \textit{phase recognition}. Movements were sampled at $100$ Hz. The number of features was set to $N=9$, with parameters, in Eq. \ref{feateq}, $c_k=(k-1)/(N-1)$ and $h_k=0.15$.

\section{Results}
\begin{figure}[htbp]
	\centering
		\includegraphics[width=0.90\columnwidth]{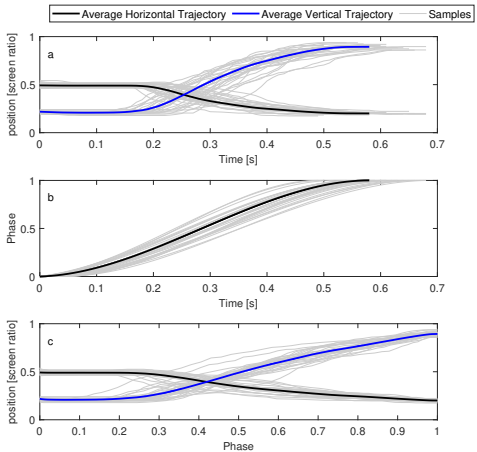}
	\caption{Dataset "mov 1" (see Fig. \ref{dataset}). (a) shows vertical and horizontal coordinates over time, gray lines represent training data, black line the average. In (b) the corresponding phase profiles are shown, in gray the ones associated with the observations, in black the average one. In (c) the resulting movements over phase are shown.}
	\label{fig:Fig3Fiteps}
\end{figure}

\begin{figure}[htbp]
	\centering
		\includegraphics[width=0.90\columnwidth]{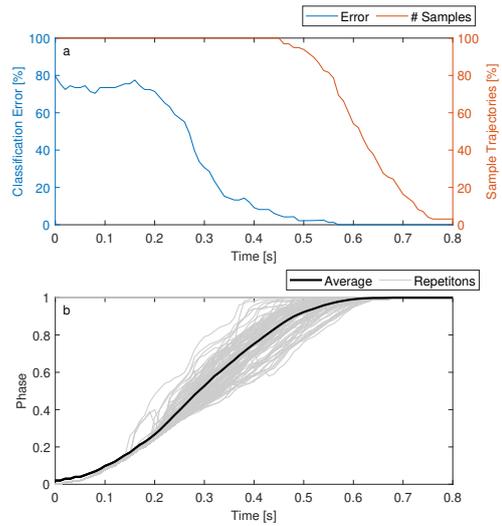}
	\caption{Movement and phase recognition accuracy evaluated on a test set of 100 samples (25 for each of the 4 recognized movements). In (a) the relative error is shown as a function of time (blue). As the different test samples have different durations. The number of available movement samples compared at a given time is shown (orange). The classification error decreases from random (75\%) as more of the sample movement is observed. In (b) the estimated phases are shown (gray). Notice that the likelihood of every phase value is computed at each point in time using Eq. \ref{phaseid} and the most likely phase is selected. The resulting phase signal differs from the ones assumed to generate the training set (Eq. \ref{betaeq}), and in general is not guaranteed to be monotonic. Nevertheless, the average phase (black) recovers the typical sigmoidal profile.}
	\label{fig:ResultsLabelPhase}
\end{figure}

\paragraph{Model identification, movement recognition and phase recognition.} The distribution of phase profiles obtained with Eq. \ref{mina}, Eq. \ref{minb} is shown in Fig \ref{fig:Fig3Fiteps} (b). Each sample is transformed in an MP, i.e. the parameters $(w,\phi)$ are extracted, $4$ \textit{ProMP} models representing the $4$ movements are estimated. The observed movement is identified as belonging to the most likely model to produce it according to Eq. \ref{movid}. The movement recognition is performed at each time step. The classification accuracy increases with time as shown in Fig. \ref{fig:ResultsLabelPhase} (a), and reaches $100\%$ at the end of the movement when movements are unambiguously differentiated by their position. Phase recognition is also performed for each time step using Eq. \ref{phaseid}. The phase estimate is obtained sample-by-sample and need not be monotonic as shown in \ref{fig:ResultsLabelPhase} (b). 

\paragraph{Movement Generation.}

\begin{figure*}[htb]
	\centering
	
	\begin{tabular}{cc}
		\includegraphics[width=1.00\columnwidth]{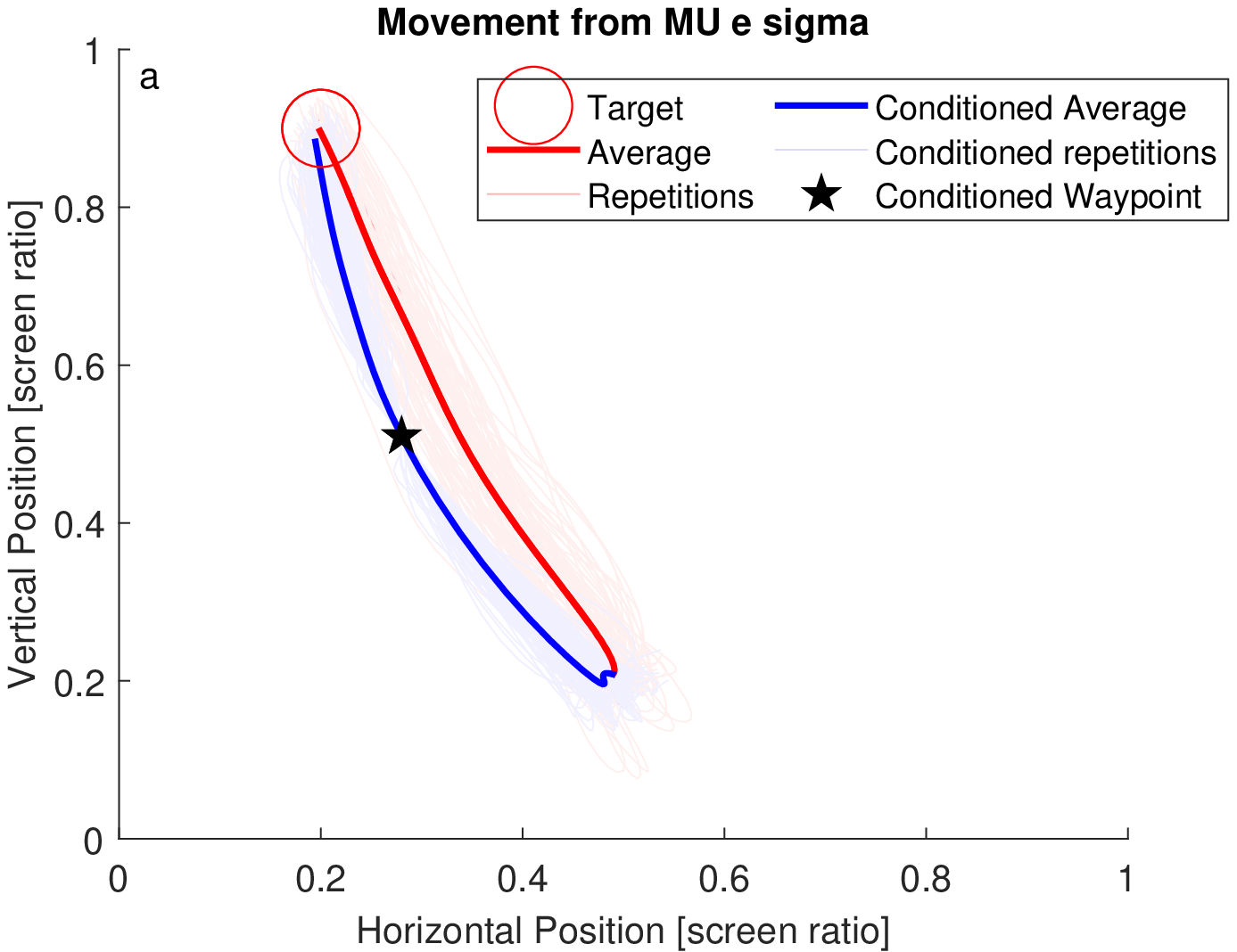} &
		\includegraphics[width=1.00\columnwidth]{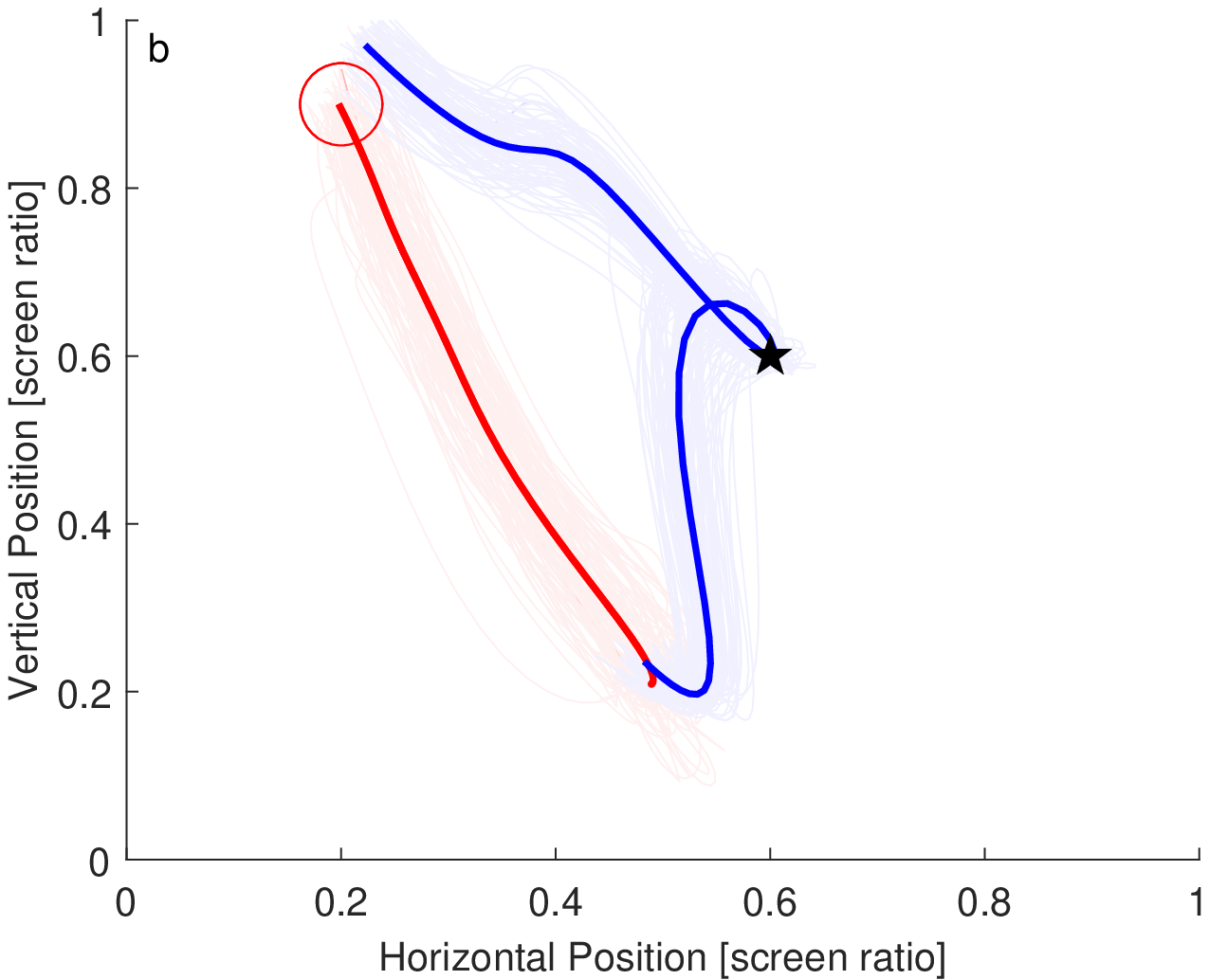} \\
		\includegraphics[width=1.00\columnwidth]{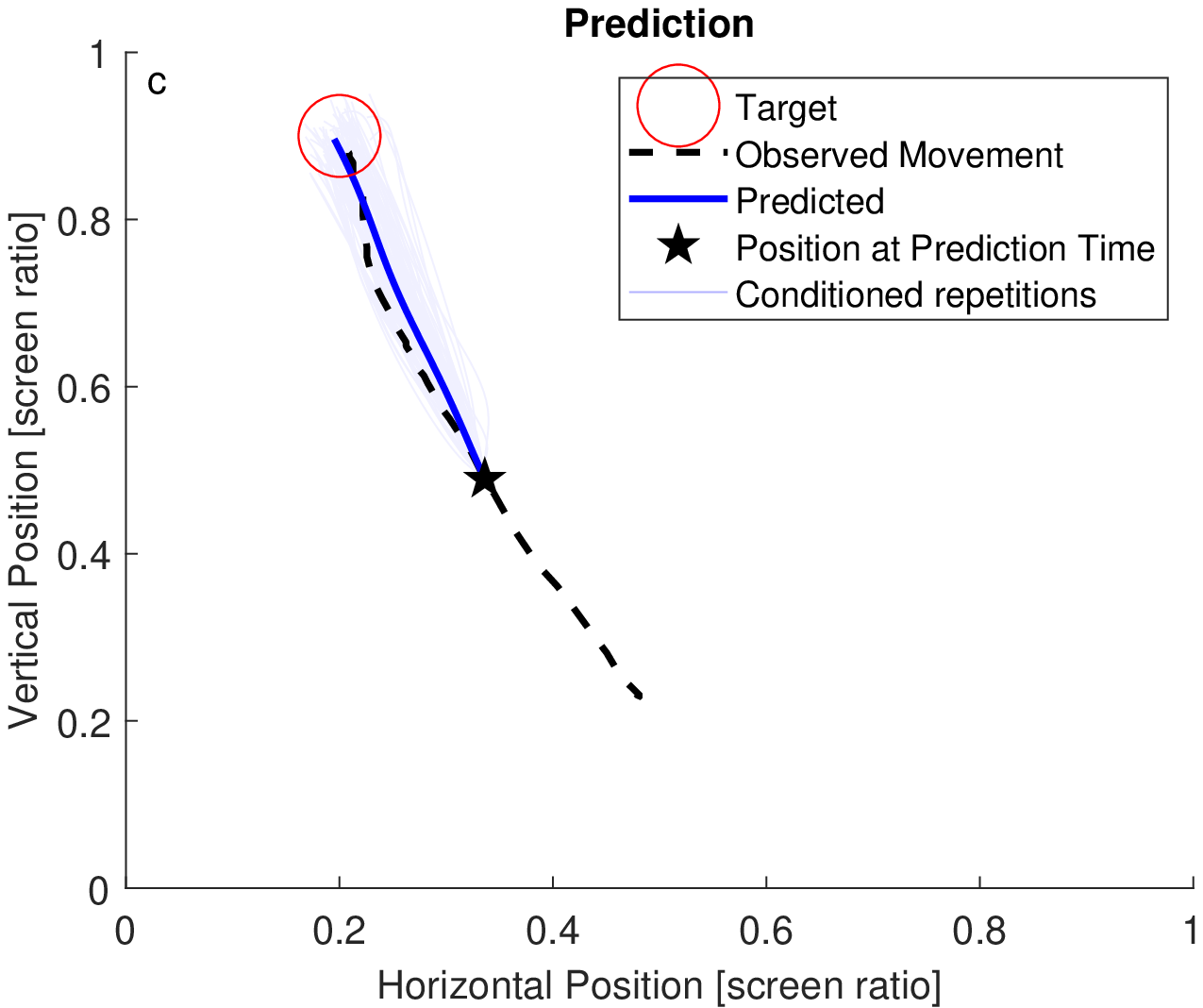} &
		\includegraphics[width=1.00\columnwidth]{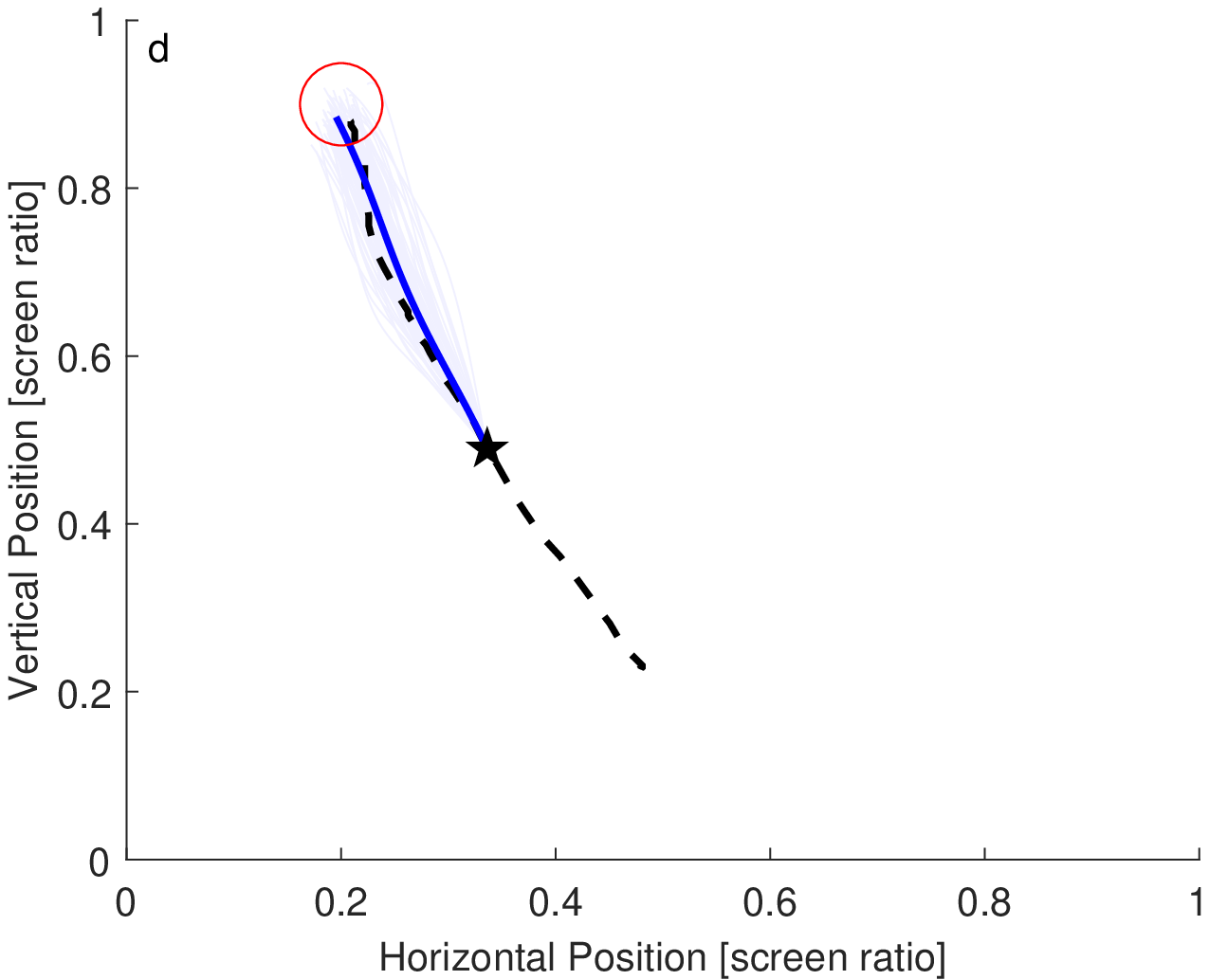} \\
		\end{tabular}	
			
	\caption{Movements generated using the ProMP model for the observations set "mov 1" (see Fig. \ref{dataset}). Trajectories in red are samples from the distribution over $w$. Trajectories in blue are samples constrained to pass through a via-point at phase $\phi^*=0.5$, indicated by a black star. Thick lines indicate the average trajectories of the respective distributions. The red circle marks the final target. In (a) the via-point is within the space of the presented examples and hence the trajectories resemble thoses presented in the training set. In (b) the via-point is an outlier to the positions presented in the training set and hence the conditioned trajectories are not representative of the observed movements. In (c) the conditioned distribution is used together with classification and phase identification to produce a prediction of the motion in progress (black dotted line). The most likely trajectory is plotted with the blue thick line, and samples as thin blue lines. In (d) the prediction is performed using a neural network to estimate the phase.}
	\label{fig:Generated}
\end{figure*}

In order to generate concrete movements the parameters $w$ are sampled from the distribution. Fig. \ref{fig:Generated} shows movements  generated with and without constraints. The constraint is expressed as a via-point to be reached at a given phase. The constrained movements are generated from the conditioned distribution with parameters from Eq. \ref{munew}, Eq. \ref{sigmanew}. In Fig. \ref{fig:Generated}(a) the imposed via-point is within the space of the presented examples and hence trajectories resemble those presented in the training set. In Fig. \ref{fig:Generated} (b) the via-point is more distant from the positions presented in the training set and hence the conditioned trajectories are less representative of the original distribution, e.g. forming loops and missing the final target. In (c) the conditioned distribution is used to predict a continuation of the current motion. The prediction is also affected by the uncertainty about the current phase, which is estimated according to Eq. \ref{phaseid}. 

\begin{figure}[htbp]
	\centering
	\includegraphics[width=0.90\columnwidth]{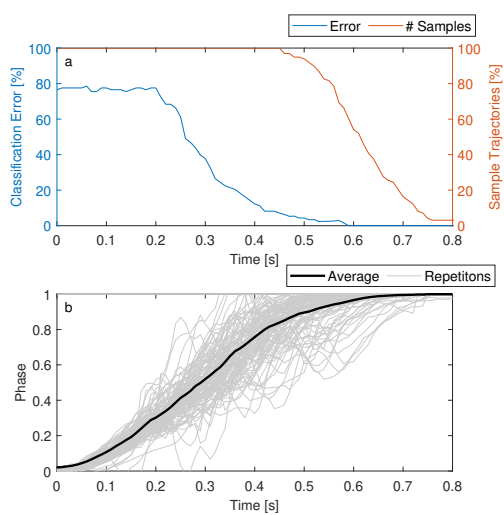}
	\caption{Classification and phase evaluation accuracy using a neural network for the identification of phase. The training set and the test set are the same as used in Fig. \ref{fig:ResultsLabelPhase}. In (a) the relative error is shown as a function of time (blue). As the movements have different durations, the number of available movement samples at a given time is indicated in orange. The classification accuracy increases as more of the test movements are observed. In (b) the identified phase profiles are shown in gray, where phase is computed at each time step. The average phase (black) shows the typical sigmoidal profile. Although in some cases the phase appears less accurate using the neural network estimator, overall classification performance is comparable to that of Fig. \ref{fig:ResultsLabelPhase}. }
	\label{fig:Resultsperc}
\end{figure}

\paragraph{Perception.} Perceptual estimation of phase was implemented with a feedforward neural network with three layers with 40, 20 and 10 neurons respectively, taking as input the current time and a vector of 20 previous samples. The network has been trained with the Levenberg-Marquardt backpropagation algorithm\cite{levenberg1944method,marquardt1963algorithm}. The results are shown in Fig. \ref{fig:Resultsperc} where the classification task (shown in Fig. \ref{fig:Resultsperc}) is repeated by exploiting the estimated phase, according to Eq. \ref{percphase}. Classification performance using the neural network is comparable to that obtained by integrating the phase distribution (Eq. \ref{movid}). In Fig. \ref{fig:Generated} (d) the phase estimated by the neural network is used to perform a prediction on the basis of the observed trajectory. The result is similar to that which is shown in Fig. \ref{fig:Generated} (c).  

\paragraph{A 7-DoF robot arm example.} In additon to previous low-dimensional examples, model identification is applied to movements of a 7-DoF robotic arm. 

\begin{figure}[htbp]
	\centering
		\includegraphics[width=0.70\columnwidth]{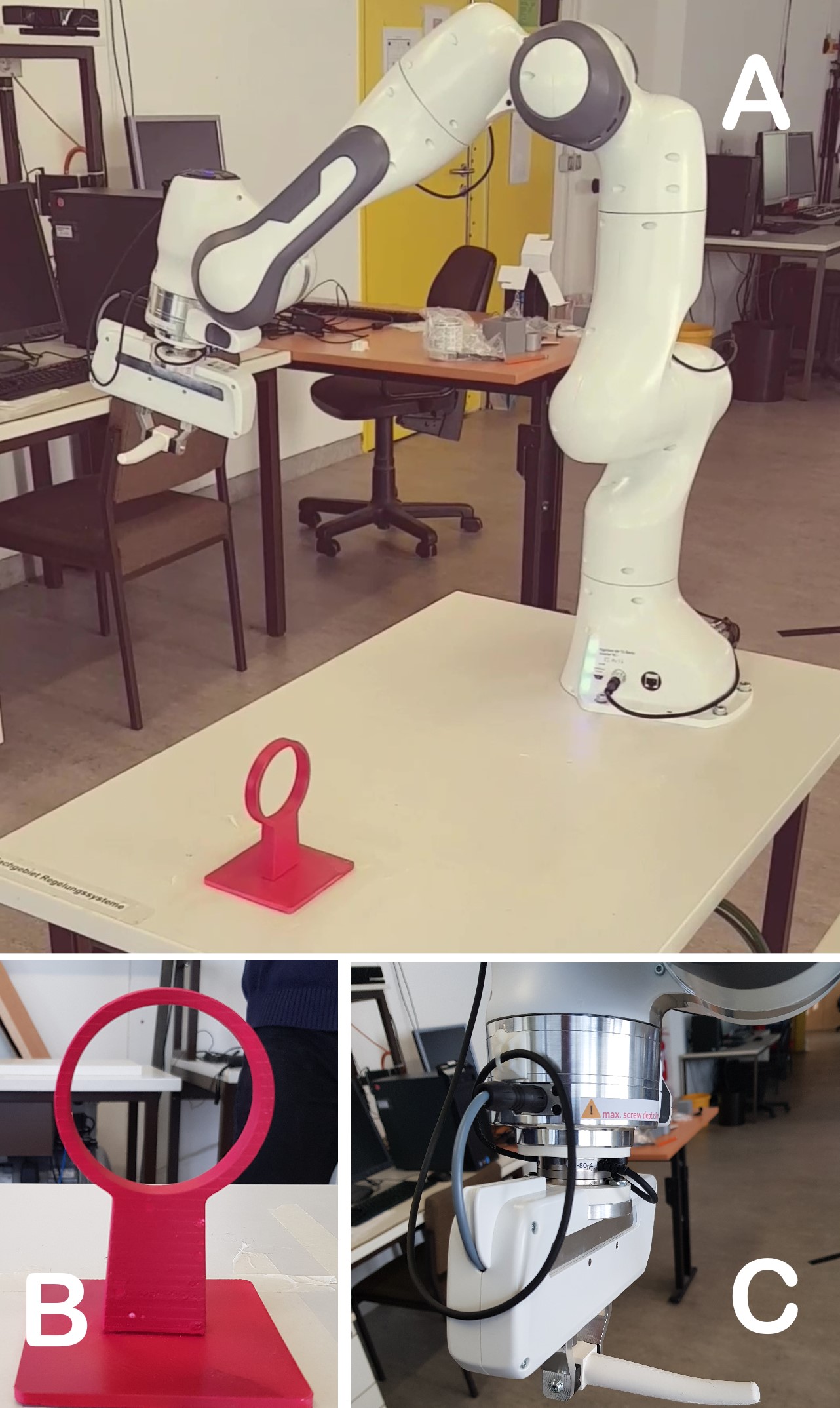}
	\caption{The robot task configuration (A). The robot's gripper is holding a hook (B) which it uses to pick up the object (C).}
	\label{robot}
\end{figure}

\begin{figure}[htbp]
	\centering
		\begin{tabular}{cc}
		\includegraphics[height=0.30\columnwidth]{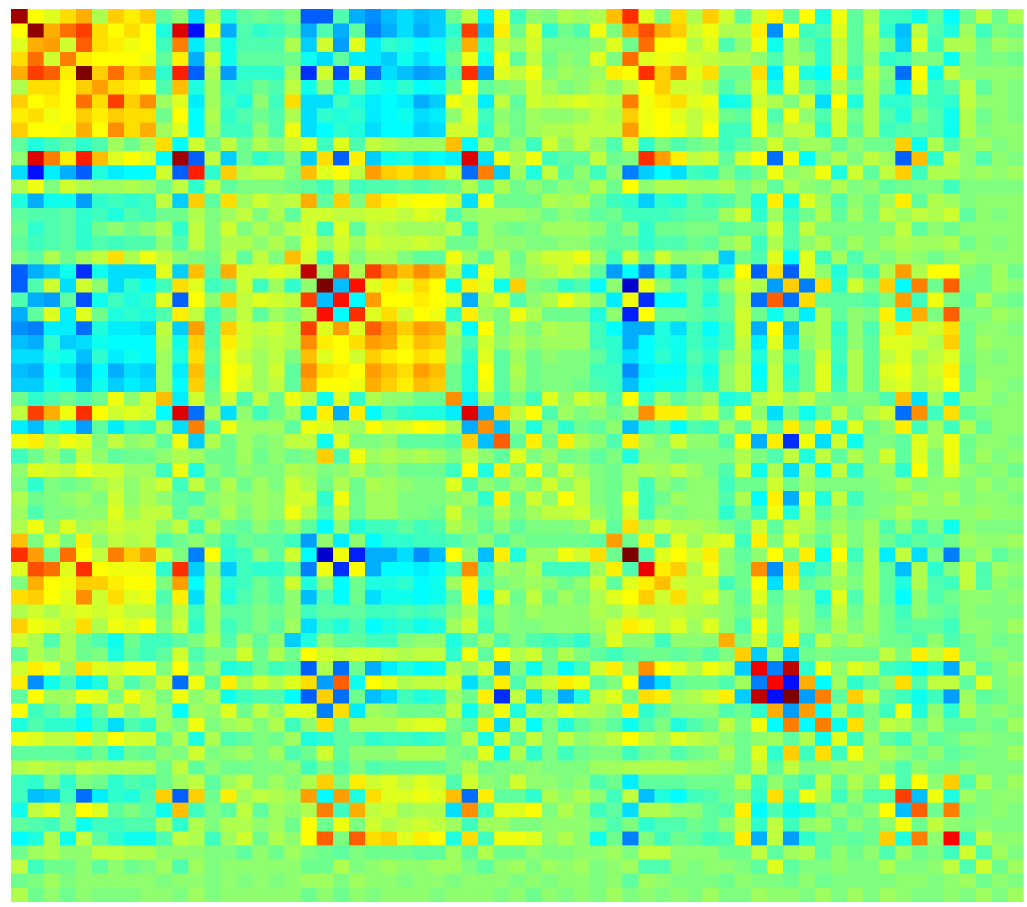}	&	
		\includegraphics[height=0.30\columnwidth]{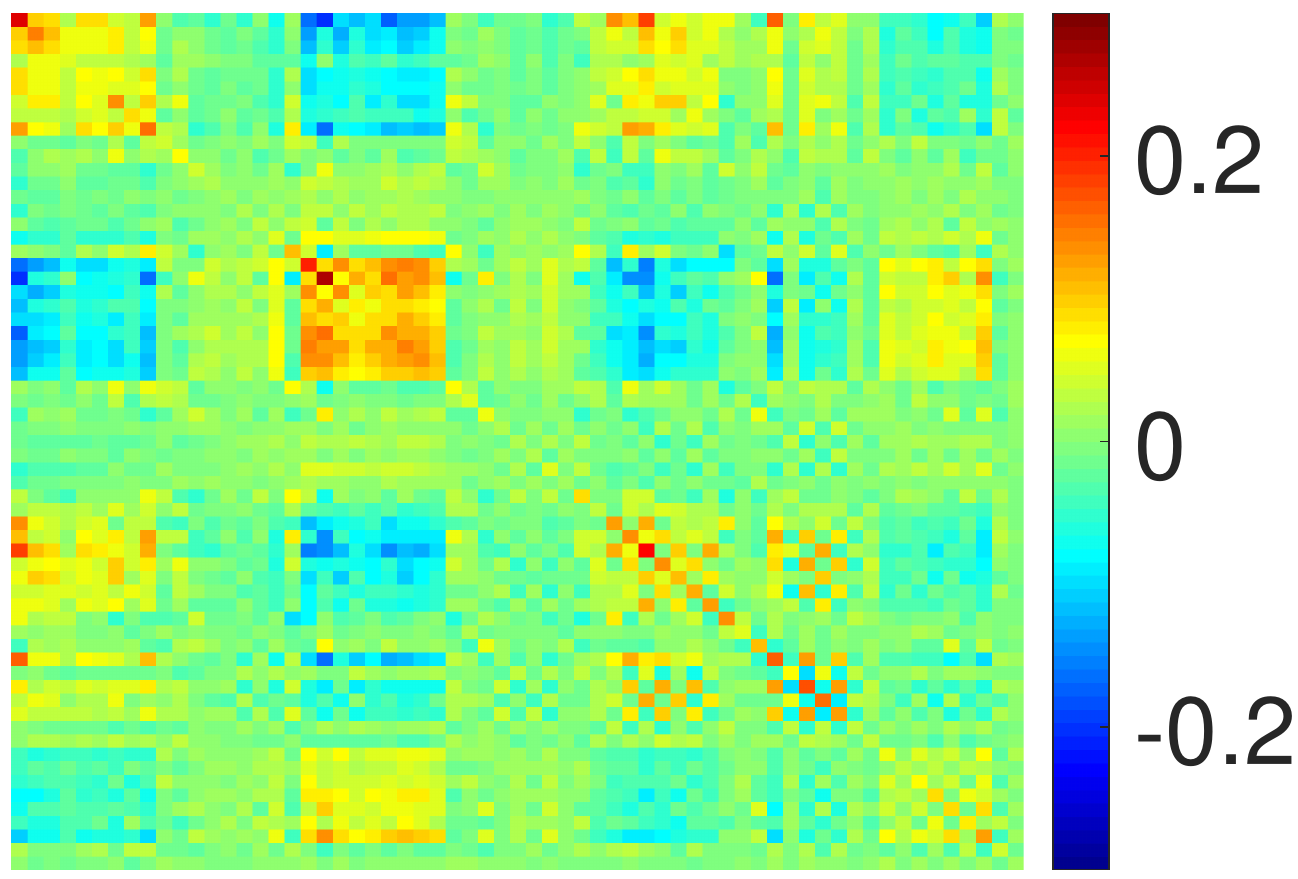}
		\end{tabular}
		
	\caption{Color representation of the covariance matrix for $w$, for the robot arm movements set without phase modulation (left) and with phase scaling (right). Most of the values in the covariance matrix are smaller.}
	\label{covmat}

\end{figure}

\begin{figure}[htbp]
	\centering
		\includegraphics[width=0.90\columnwidth]{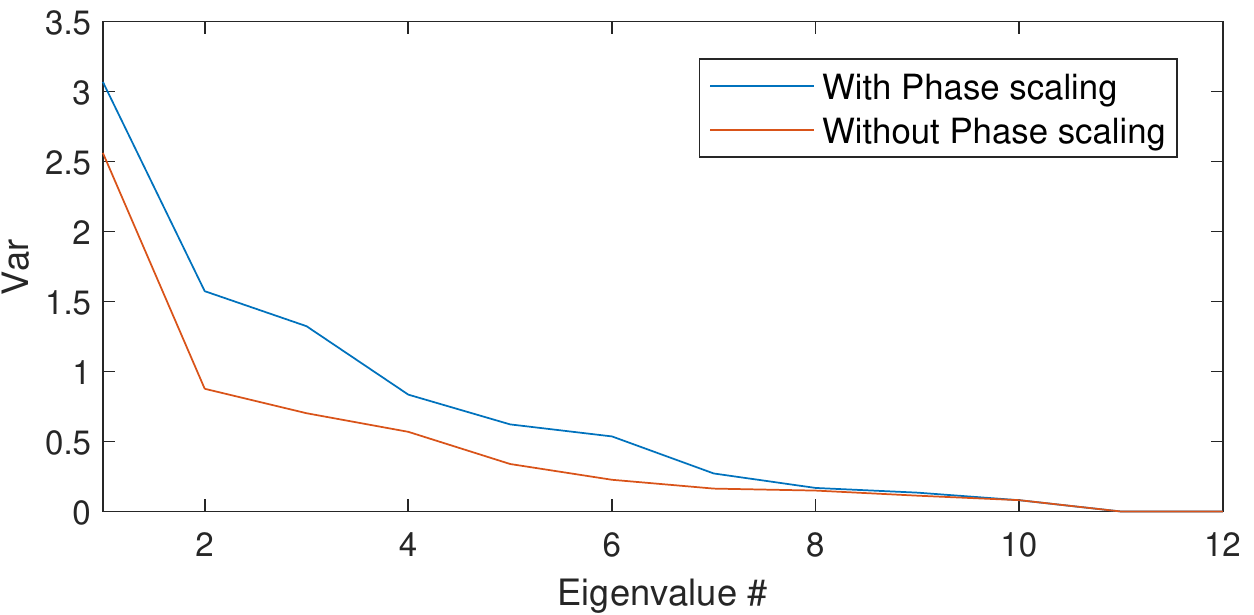}
	\caption{Non-zero eigenvalues of the covariance matrices shown in Fig. \ref{covmat}. With the phase modulation all eigenvalues are smaller, which indicates a more precise reproduction of movements.}
	\label{fig:Eigenv_Crop}
\end{figure}

The task consists in picking up an object using a hook and handing it to an user in a single continuous movement. The setup is shown in Fig. \ref{robot}. The object's loop represents a spatial constraint for the trajectory, since the object is always picked up in the same position. Trajectory examples were recorded by manually moving the robot, producing a total of $11$ examples. The task was represented in seven dimensional joint space. The number of features was set to $9$, hence the ProMP distribution has $63$ dimensions.
Fig. \ref{covmat}. shows the covariance matrices of the ProMP's distributions learned on data without and with optimized phase profiles. 
The use of phase modulation decreases variation in adjacent feature weights in the covariance matrix elements as temporal noise is externalized to the phase signal. A more compact description of the concept is shown in Fig. \ref{fig:Eigenv_Crop}, where  the eigenvalues of the two covariance matrices, i.e. the \textit{principal components}, are compared. All the non-zero eigenvalues are smaller with phase modulation.



\section{Discussion and Future Work}
In this work we used ProMPs to model human movements and demonstrations to robots in order to perform movement recognition and movement generation, with particular emphasis on leveraging a parametric phase profile to modulate movement execution speed. The approach showed to properly classify and generalize the movements in a data-set of reaching movements. The proposed parametric phase profile has a \textit{sigmoidal} shape that is particularly suitable to reaching tasks such as the one presented into the example and other "real world" tasks, such as pick and place operations where it is relevant to the control of initial and final position. However, a sigmoidal phase profile enforces zero velocity at the beginning and at the end of the movement. Therefore, in tasks where final velocity must be controlled, other profiles (e.g. a beta function with $a=2$, $b=1$) may be favorable. In general it should be considered that time scaling of human movements can work differently depending on the task and the working conditions\cite{zhang1999effects}. The example reaching task is represented as a library of $4$ ProMPs describing movements to $4$ possible targets. In general, the diversity of movements in a task can be represented with a manifold in the space of hyper-parameters and mixture distributions \cite{rueckert2015extracting}, or specifically with gaussian mixtures describing the distribution of $w$ \cite{ewerton2015modeling}. The parameters can be constrained with linear relationships for dimensionality reduction, as recently shown in \cite{colome2018dimensionality}. With human movements, dependencies between target positions and the moving parameters can be represented by a linear model\cite{avizzano2011regression,lippi2012method}. In all those cases phase can be estimated online while observing the movement by modifying Eq.~\ref{phaseid} to take into account the chosen distribution. Further, a specific perceptual estimator dedicated to the identification of phase using a feed-forward neural network was described and implemented. The estimator led to a performance in terms of classification and prediction accuracy that was comparable to that of the system exploiting the empirical probability distribution of the phase directly. An advantage of using a neural network is, that it avoids the potentially costly computation of the integral in Eq.\ref{movid} at each integration step. Future work on phase recognition will be focus on the identification of state of dynamic system models tasked with generating phases, e.g. \textit{phase-state machine} models \cite{deimel_dynamical_2019}. Such models could provide phase profiles and could be used in conjunction with a library of ProMPs to control a task. Furthermore, current research on ProMPs includes the design of feedback control systems considering physical interaction with the environment and the users  \cite{paraschos2018using,paraschos2013probabilistic}. Time scaling does not apply to the control of arm kinematics and the external contact forces in the same way as for movements (this holds for each kind of rescaling, also the linear one). A general framework that can scale kinematics and contact instants in time should be integrated with a consistent control of applied forces and torques.

\balance 

\section*{\uppercase{Acknowledgements}}
We gratefully acknowledge financial support for the project MTI-engAge (16SV7109) by BMBF

\bibliographystyle{apalike}
{\small
\bibliography{lippi}}


\vfill
\end{document}